\documentclass[journal]{IEEEtran}
\usepackage{graphicx}
\usepackage{amsmath,amsfonts,amssymb}
\usepackage{algorithmic}
\usepackage{array}
\usepackage{multirow}
\usepackage{amsmath}
\usepackage{subfig}
\usepackage{url}
\usepackage[pdfstartview=FitH,
            CJKbookmarks=true,
            bookmarksnumbered=true,
            bookmarksopen=true,
            linkcolor=red,
            anchorcolor=red,
            citecolor=green
            ]{hyperref}
\usepackage[ruled,vlined]{algorithm2e}

\SetKwInput{HyperPara}{Hyper-Para}
\SetKwBlock{TrainStage}{Training Stage}{End Train Stage}
\SetKwBlock{InferStage}{Inference Stage}{End Infer Stage}
\SetKwBlock{FgFMoE}{Fine-grained Texture MoE Layer}{End Fine-grained Texture MoE}
\SetKwBlock{CgSMoE}{Coarse-grained Structure MoE Layer}{Coarse-frained Structure MoE Layer}
\SetKwBlock{SparseAc}{TopK Expert Activation}{End TopK Expert Activation}
\SetKwIF{AnnealFull}{AnnealSparse}{Anneal Branch}{Full Expert Activation}{}{End Anneal Branch}

\begin{document}

\title{Mutual Modality Trust with Lightweight Reconstruction Regularization for Fine-grained Tire Pattern Recognition}

\author{\IEEEauthorblockN{Jianning Yang}, Jie Fang$^{\dag}$, Xinda Ma, Zirui Song, Dianwei Wang, and Nan Wang

\thanks{J. Yang, J. Fang, X. Ma, Z. Song and D. Wang are with the School of Communications and Information Engineering, Xi'an University of Posts and Telecommunications, Xi'an 710121, China. N. Wang is with the School of Artificial Intelligence, Hainan Normal University, Haikou 571158, China. J. Fang is also with the School of Artificial Intelligence, Optics and Electronics (iOPEN), Northwestern Polytechnical University, Xi'an 710072, China.

$^{\dag}$Correspondence (Email: fangjie@xupt.edu.cn)}}

\maketitle

\begin{abstract}
Visual tire recognition serves as a core supporting technique for vehicle safety monitoring, autonomous driving perception and automated automotive maintenance. Existing fine-grained tire recognition techniques suffer from three prominent limitations. They tend to depend on only one visual source, lack the capacity to jointly model spatial and frequency cues for minute tread texture extraction, and suffer severe overfitting given limited annotated tire imagery. This paper proposes a lightweight fine-grained tire pattern recognition method incorporating dual-branch independent inference and enhanced feature fusion to boost recognition performance. The framework employs two task-specialized branches dedicated to tire surface and tread indentation, respectively, to extract modality-specific discriminative features. Each branch conducts independent prediction, while cross-branch feature fusion exploits Mutual Modality Trust (M$^2$T) to realize complementary feature enhancement across two modalities. Besides, a frequency-domain hierarchical guidance module is devised, which leverages bandpass filters to decompose feature maps into high- and low-frequency components and enables fine-grained cross-layer feature modulation. Furthermore, a Lightweight Reconstruction Regularization (LR$^2$) is introduced to retain abundant intrinsic information within feature embeddings, substantially improving feature stability and recognition robustness under limited labeled training data. In addition, we establish a surface-indentation multi-source dataset namely MTire299 for fine-grained tire tread recognition, which covers 299 categories with a total of 14795 paired image samples. Extensive experiments conducted on two public tire datasets validate the superiority and efficacy of the proposed algorithm.
\end{abstract}


\section{Introduction}
Tires constitute the only vehicle component that maintains direct contact with pavement. Their tread layouts, structural properties and wear states fundamentally determine driving stability, braking reliability and steering controllability, laying an essential foundation for overall road traffic safety. Driven by the booming development of intelligent transportation, autonomous driving and automotive aftermarket services, automated high-precision tire identification has emerged as an indispensable technical pillar for industrial intelligent transformation. Reliable recognition of tire specifications, tread textures and surface defects delivers credible data support for onboard vehicle monitoring and adaptive driving control, while also enabling proactive prediction of potential traffic hazards. Meanwhile, fine-grained tire identification streamlines tire matching and wear quantification across waste tire recycling, vehicle maintenance and insurance damage appraisal workflows, greatly lifting operational efficiency and intelligent management standards within related industries. Real-world tire inspection tasks still face prominent practical obstacles. Mass tire categories share nearly identical visual appearances, which raises the difficulty of fine-grained discrimination. Conventional manual inspection suffers from low throughput and strong subjective bias, frequently triggering misjudgments that fail to meet the demands of large-scale, standardized industrial detection.

\begin{figure}[t]
	\centering
	\includegraphics[width=1.0\linewidth]{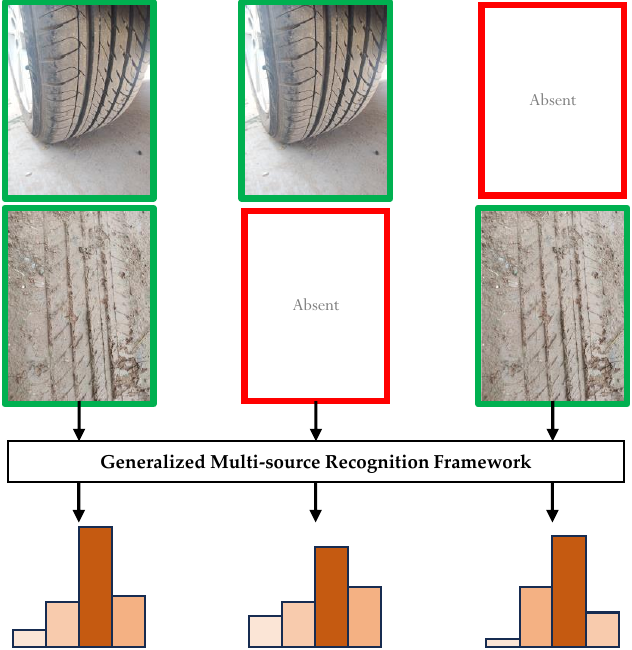}
	\caption{Inference under various modal combinations of our proposed method.}\label{fig:motivation}
\end{figure}

In these cases, a series of vision-based fine-grained tire pattern recognition have been proposed successively because of its convenience and objectivity. For instance, Liu et al. \cite{liu2019rotation} proposed a rotation invariant HOG descriptor to extract features that robust to illumination, scale changes, and rotation invariant to improve the recognition performance. Yan et al. \cite{yan2021tire} proposed a few-shot learning recognition framework based on classical CNN features, which calibrates the distribution of few sample classes through transferring statistics from the classes with sufficient features. Feng et al. \cite{feng2023fine} incorporated knowledge distillation and attention mechanism into the deep neural network to enhance its representation capability, and further improve the recognition performance.

However, tire images captured under complicated real-world conditions are disturbed by surface stains, occlusion, illumination fluctuations, viewpoint shifts, tread wear and deformation. Subtle intra-class variations within identical tire categories are hard to discriminate. The challenges are further exacerbated by small-scale existing tire datasets, scarce annotated samples and severe confusion of fine-grained features. Conventional recognition approaches suffer from weak feature extraction and insufficient robustness, which cannot support high-precision tire identification in complex environments. Current deep learning-based methods have gained certain performance gains, yet their overall performance still cannot match practical application standards. The major bottlenecks are outlined below.

\begin{enumerate}
         \item Few methods jointly exploit tire surface textures and indentations, instead relying on a single cue for recognition and yielding degraded performance.
         \item Traditional CNNs rely on massive training data for model optimization, yet fine-grained tire pattern recognition datasets are usually limited in scale.
	\item SOTA neural networks perform spatial-domain encoding on tire images, whose extracted representations cannot adequately reflect statistical texture features.
\end{enumerate}

To address the abovementioned issues, we propose a generalized multi-source fine-grained tire pattern recognition algorithm, which can address recognition problem under different modal combinations as shown in Fig. \ref{fig:motivation}. Specifically, it employs two identical-architecture ConvNeXt-Tiny branches dedicated to tire surface and tread indentation to extract modality-specific discriminative features, and each branch conducts independent prediction, while cross-branch feature fusion exploits M$^2$T to realize complementary feature enhancement across two modalities. Besides, a frequency-domain hierarchical guidance module leverages bandpass filters to decompose feature maps into high- and low-frequency components and enables fine-grained cross-layer feature modulation. Furthermore, a LR$^2$ is introduced to retain abundant intrinsic information within feature embeddings, substantially improving feature stability and recognition robustness under limited labeled training data. In addition, we establish a surface-indentation multi-source dataset namely MTire299, which covers 299 categories with a total of 18,233 paired image samples, to address the insufficient scale of existing datasets. Overall, the main contributions of the proposed method can be summarized as follows. 
\begin{itemize}
	\item We design a dual-branch fine-grained tire pattern recognition framework toward M$^2$T capable of inference from separate branches or the fused branch as required.
         \item We adopt MobileConvNeXt-Tiny as the backbone of each branch and introduce a LR$^2$ to constrain the optimization process to mitigate overfitting risks.
	\item We propose a hierarchical bandpass filtering-guided cross-domain weighting strategy for each modal encoder to boost collaborative spatial-frequency representations.
          \item We establish a large-scale of surface-indentation multi-source dataset, to promote the development of fine-grained tire pattern recognition.
\end{itemize}

The rest of this paper is structured as follows. Sec. \ref{sec:relatedworks} reviews the existing fine-grained tire pattern recognition methods briefly. Sec. \ref{sec:algorithm} introduces the proposed algorithm in detail. Sec. \ref{sec:experiments} reports the experiments comprehensively, and Sec. \ref{sec:conclusion} concludes the paper.

\section{Related Works}\label{sec:relatedworks}
Existing tire pattern recognition approaches fall into three mainstream categories: handcrafted feature-based algorithms, vanilla convolutional neural networks (CNNs), and sophisticated neural networks. Handcrafted feature methods enjoy the merits of low computational overhead and high interpretability, yet they fail to capture discriminative fine-grained tire patterns. Although vanilla CNNs enable end-to-end automatic feature extraction, they are deficient in multi-modal feature interaction and exhibit weak generalization against complex real-world scenarios. In contrast, sophisticated neural networks integrate well-designed multi-branch architectures, attention modules and knowledge distillation schemes, thus delivering prominent improvements on fine-grained tire recognition tasks.

\subsection{Handcrafted feature methods}
Research on handcrafted features for tire pattern recognition can be categorized into shape, texture, and multiple feature extraction paradigms, as reviewed below.

In the branch of shape feature representation, Hu et al. \cite{hu1962visual} pioneered the invariant moment theory, which generates translation-, rotation-, and scale-invariant moment descriptors and has long served as a fundamental tool for early object matching and tire contour characterization. To further capture local geometric cues, Dalal and Triggs \cite{dalal2005histograms} developed the HOG framework to encode spatial edge and gradient statistics; this descriptor is commonly fused with other handcrafted features to facilitate tire tread contour detection tasks.

For texture characterization, Ojala et al. \cite{ojala2002multiresolution} proposed the classical LBP operator, an efficient lightweight descriptor that models local gray-level variations and dominates mainstream tire surface texture analysis pipelines. Complementing spatial-domain handcrafted features, frequency-domain transformation techniques provide an alternative way to decouple tire image components. Specifically, Mallat \cite{mallat1989theory} established the complete theoretical framework for multi-resolution wavelet decomposition, enabling separate extraction of low-frequency contour components and high-frequency fine details from tire images. Building upon multi-scale frequency filtering, Zhang et al. \cite{muthohhar2024classification} exploited Gabor filter banks to extract multi-scale frequency features, realizing automated grading of tire tread wear with promising performance under real traffic surveillance scenarios.

Beyond single-type feature extraction, recent studies have explored the fusion of multiple handcrafted features to strengthen discriminability. Nevertheless, Li et al. \cite{hongling2023feature} demonstrated that naive feature concatenation fails to fully exploit the complementary latent information embedded in tire images, revealing inherent limitations of straightforward fusion strategies. Aiming at practical tire inspection tasks, several improved handcrafted pipelines have been devised. Li et al. \cite{li2024tire} optimized conventional image operators and multi-feature combinations to accelerate tire surface damage identification. Liu et al. \cite{liu2023tire} integrated HOG with a refined LBP variant for tire appearance defect detection, moderately boosting the noise robustness of handcrafted feature representations. Extending the applicable scenarios of handcrafted schemes, Poquet et al. \cite{poquet2013tyre} integrated image alignment with handcrafted attribute descriptors to accomplish precise matching of tire prints with diverse specifications, broadening the utilization of handcrafted features in vehicle traceability applications.

\subsection{Vanilla CNNs methods}
As fundamental deep learning backbones, vanilla CNNs hierarchically extract visual representations directly from raw input images, naturally overcoming the inherent drawbacks of handcrafted feature pipelines. Constructed by cascaded convolution, pooling and fully connected layers, vanilla CNNs support end-to-end feature learning and classification without labor-intensive manual feature engineering. Nevertheless, such basic architectures exhibit prominent drawbacks when deployed for tire analysis. They are prone to overfitting and deliver degraded robustness given limited training samples or noisy tire imagery.

The technical paradigm of vanilla CNNs was originally established by LeCun et al. \cite{lecun1998gradient}. The authors developed LeNet-5 as the first functional CNN, which combines convolution and pooling operators for handwritten digit classification. CNN research remained stagnant for decades afterward until the landmark AlexNet proposed by Krizhevsky et al. \cite{krizhevsky2012imagenet}. Equipped with ReLU activation, dropout regularization and GPU parallel acceleration, AlexNet attained leading accuracy on the ImageNet benchmark and reignited widespread research interest in convolutional networks. Building on this breakthrough, Simonyan and Zisserman \cite{simonyan2014very} presented VGGNet. This network stacks successive 3×3 convolutional kernels to increase network depth, and the strengthened feature discriminability renders VGGNet a standard baseline for vanilla CNN-based vision tasks.

Despite the solid theoretical framework of vanilla CNNs, their performance on tire inspection tasks is constrained by insufficient data and noise interference. A series of regularization and normalization techniques have been proposed to mitigate these issues. Srivastava et al. \cite{srivastava2014dropout} devised the dropout strategy to suppress overfitting. This lightweight module has become a standard component for vanilla CNNs trained on small-scale tire datasets. Further, Ioffe and Szegedy \cite{ioffe2015batch} put forward batch normalization to stabilize the training distribution of deep vanilla CNNs, speeding up model convergence for tire defect identification. On the practical application side, Tada et al. \cite{tada2017defect} deployed a shallow vanilla CNN to categorize internal tire surface flaws, verifying the feasibility of basic CNN architectures for industrial rubber defect inspection.

\subsection{Sophisticated NNs methods}
To remedy the inherent limitations of vanilla convolutional neural networks (CNNs) in tire pattern analysis and defect detection tasks, advanced neural network architectures have been developed with elaborate structural designs, including residual connections, dense feature propagation, adaptive attention modules, and transformer paradigms. These advanced networks substantially strengthen the capability of learning fine-grained textures, multi-scale contextual information, and long-range feature dependencies, thus achieving robust tire feature extraction against complex industrial disturbances such as uneven illumination, environmental noise, and target occlusion. However, the boosted representation performance comes at the expense of higher computational overhead, and these sophisticated models are prone to overfitting on limited tire datasets, necessitating reasonable regularization strategies in practical deployment.
\begin{figure*}[t]
	\centering
	\includegraphics[width=1.0\linewidth]{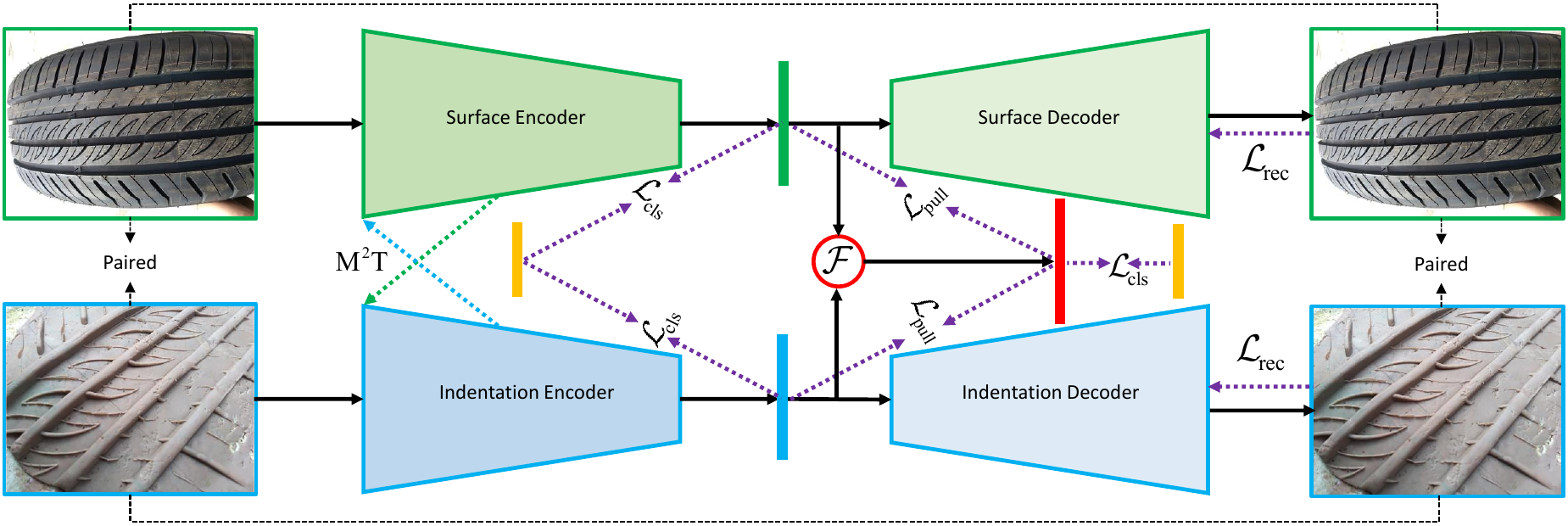}
	\caption{The overview framework of the proposed method.}\label{fig:method}
\end{figure*}

To further enhance the feature discrimination of CNNs for tiny and ambiguous tire defects, attention mechanisms and vision transformers have been introduced to adaptively recalibrate feature weights and model global spatial dependencies. Early attention research focuses on channel feature optimization: SENet \cite{hu2018squeeze} performs channel-wise feature recalibration to suppress redundant features and highlight subtle defect information on tire surfaces. To complement spatial modeling capability, CBAM \cite{woo2018cbam} integrates channel and spatial attention modules to hierarchically refine tire tread feature representations, further improving the model’s anti-interference ability. Beyond the local receptive field constraint of CNN-based methods, Vision Transformer (ViT) \cite{dosovitskiy2020image} exploits self-attention mechanisms to capture global long-range dependencies, providing a new perspective for comprehensive tire pattern modeling. Combining the complementary strengths of CNNs, attention modules, and transformers, hybrid detection models have been widely applied to high-precision tire defect tasks. Peng et al. \cite{peng2023td} embedded attention mechanisms into the YOLO framework and proposed TD-YOLOA, realizing precise detection of tiny defects in tire radiographic images. Yang et al. \cite{yang2022tire} fused ViT and residual networks to simultaneously extract local texture and global structural features, effectively improving the accuracy of tire wear classification. For complex tire recycling scenarios with cluttered backgrounds, Phadnis et al. \cite{phadnis2025sustainable} constructed a ResNet50-YOLOv7 hybrid model, which achieves an excellent balance between detection accuracy and real-time inference speed.

In recent years, state-space-based ConvNeXt-Tiny models have emerged as a novel efficient visual backbone, overcoming the computational bottlenecks of CNNs and transformers and offering a superior alternative for high-efficiency visual inspection. Unlike transformers with quadratic computational complexity, ConvNeXt-Tiny series models implement global feature modeling with linear complexity via state-space theories. To adapt the original one-dimensional ConvNeXt-Tiny for visual tasks, Liu et al. \cite{liu2022convnet} designed the SS2D module to capture multi-directional spatial context, enabling effective two-dimensional visual feature extraction. To further strengthen ConvNeXt-Tiny’s spatial modeling capacity, Xin et al. \cite{xin2025multi} embedded the proposed MPSM blocks into an improved U-Net architecture, endowing ConvNeXt-Tiny with powerful fine-grained feature learning ability. Furthermore, considering the limitation of existing ConvNeXt-Tiny in long-context modeling, Lu et al. \cite{lu2026mamba} proposed a spectrum scaling strategy for transition matrices based on the model’s inherent context sensitivity, which significantly enhances the generalization capability for long-range spatial dependencies, making ConvNeXt-Tiny highly suitable for high-resolution tire defect recognition tasks.

\section{Algorithm}\label{sec:algorithm}
\subsection{Overview}
The overall paradigm of the proposed method is illustrated in Fig. \ref{fig:method}, which consists of two parallel representation branches with identical architectures for tire surfaces and tire indentations respectively, and each branch is actually an asymmetric encoder–decoder network. Besides, a bidirectional cross-modal guidance module built upon M$^2$T is introduced into two encoders to exploit complementary information between the two modalities. Furthermore, a hierarchical bandpass filtering-guided cross-domain weighting strategy is adopted for each encoder to enhance collaborative spatial-frequency representations. In addition, a LR$^2$ optimization mechanism is introduced to retain abundant intrinsic information within feature embeddings, substantially improving feature stability and recognition robustness under limited labeled training data. Finally, the fused features derived from the outputs of the two encoders are utilized for final recognition.It is noteworthy that the two separate branches and the fused branch can independently complete inference and recognition tasks, enabling the model to adapt to diverse scenarios such as missing modalities.

\subsection{M$^2$T based encoder}
\begin{figure}[htpb]
	\centering
	\includegraphics[width=1.0\linewidth]{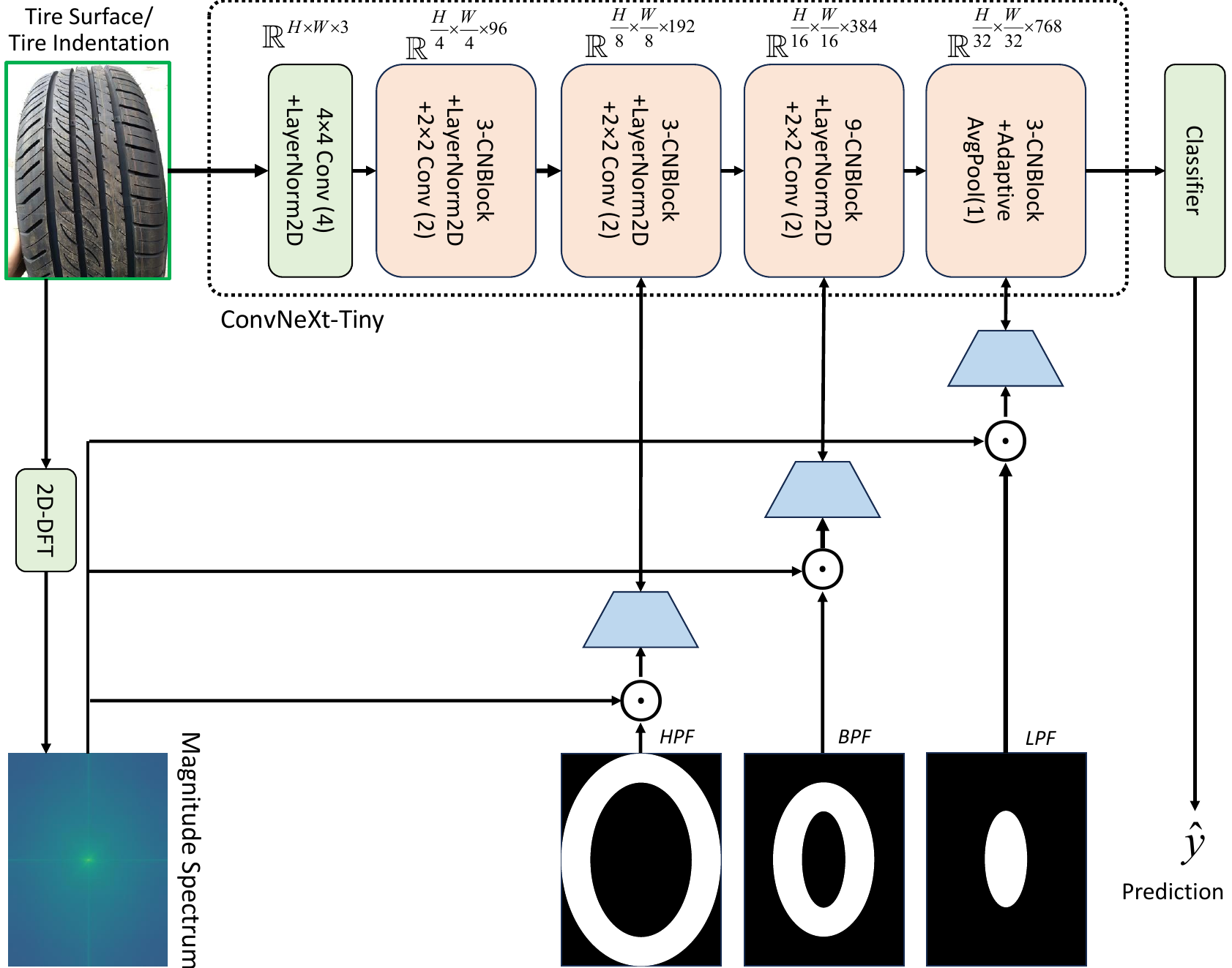}
	\caption{The inner architecture of the surface/indentation encoder based on frequency-guided MobileConvNeXt-Tiny.}\label{fig:encoder}
\end{figure}

The inner architecture of the M$^2$T based encoder is shown in Fig. \ref{fig:encoder}, which mainly contains a spatial domain flow, a frequency domain flow, and an M$^2$T module for surface-indentation cross-branch interaction.

Firstly, we utilize 2D-DFT operator to obtain frequency images corresponds to spatial ones, and its formulation is defined as Equation (\ref{eq:2ddft}).
\begin{equation}
{F_{u,v}} = \sum\limits_{w = 1}^W {\sum\limits_{h = 1}^H {{f_{w,h}} \cdot {e^{ - 2j\pi \left( {\frac{{uw}}{W} + \frac{{vh}}{H}} \right)}}} }
\label{eq:2ddft}
\end{equation}
where $W$ and $H$ denote the width and height of the image respectively. $f_{w,h}$ denotes the $(w,h)_{th}$ element of the spatial-domain image, while $F_{u,v}$ stands for the $(u,v)_{th}$ element of the frequency-domain image. Besides, we design a set of band-pass filters (BPFs) to separate different frequency components, and the $n_{th}$ BPF is defined as Equation (\ref{eq:bpf}).

\begin{equation}
\begin{array}{l}
{\rm{BP}}{{\rm{F}}_n} = Z,\\
\begin{array}{*{20}{c}}
{s.t.}&{{Z_{u,v}} = \left\{ {\begin{array}{*{20}{c}}
{1,{r_{n - 1}} \le \sqrt {{{\left( {u - \frac{W}{2}} \right)}^2} + {{\left( {v - \frac{H}{2}} \right)}^2}}  < {r_n}}\\
{0,otherwise}
\end{array}} \right.}
\end{array}
\end{array}
\label{eq:bpf}
\end{equation}
where $r_n$ denotes the $n_{th}$ cut-off frequency, in which $r_0=0, r_1=85, r_2=170, r_3=255$. Furthermore, we update the spatial feature map using Equation (\ref{eq:fsaptial}).

\begin{equation}
{{{\bf{\hat F}}}_{{\rm{spatial,}}n}} = {{\bf{F}}_{{\rm{spatial,}}n}} \odot A_{4-n}^{{\rm{F2S}}}
\label{eq:fsaptial}
\end{equation}
where $A_n^{{\rm{F2S}}}$ denotes the $n_{th}$ frequency-to-spatial guidance map, which is defined as Equation (\ref{eq:attf2s}). 

\begin{equation}
A_n^{{\rm{F2S}}} = {\bf{1}} + \sigma \left( {{\rm{FC}}{{\rm{N}}_n}\left( {{\rm{BP}}{{\rm{F}}_n} \odot F} \right)} \right)
\label{eq:attf2s}
\end{equation}
where $\sigma \left(  \cdot  \right)$ denotes the sigmoid function. ${\rm{FCN}}\left(  \cdot  \right)$ denotes the fully convolutional network operator. In addition, we utilize M$^2$T based cross-modal interaction operator to enhance the representations of both two branches, and the update process is defined in Equation (\ref{eq:mmt}).

\begin{equation}
\left[ {\begin{array}{*{20}{c}}
{{{{\bf{\hat F}}}_{{\rm{surf}},n}}}\\
{{{{\bf{\hat F}}}_{{\rm{iden}},n}}}
\end{array}} \right] = \left[ {\begin{array}{*{20}{c}}
{{\tau _{{\rm{surf}},n}}}&{1 - {\tau _{{\rm{surf}},n}}}\\
{{\tau _{{\rm{iden}},n}}}&{1 - {\tau _{{\rm{iden}},n}}}
\end{array}} \right]\left[ {\begin{array}{*{20}{c}}
{{{\bf{F}}_{{\rm{surf}},n}}}\\
{{{\bf{F}}_{{\rm{iden}},n}}}
\end{array}} \right]
\label{eq:mmt}
\end{equation}
where ${{{\bf{\hat F}}}_{{\rm{surf,}}n}}$ and ${{{\bf{\hat F}}}_{{\rm{iden,}}n}}$ denote the features of $n_{th}$ layer correspond to surface and indentation branches respectively. $\tau_{{\rm{surf}},n}$ and $\tau_{{\rm{iden}},n}$ denote the M$^2$T coefficients, and they can be obtained by Equation (\ref{eq:tauboth}).

\begin{equation}
\left[ {\begin{array}{*{20}{c}}
{{\tau _{{\rm{surf}},n}}}\\
{{\tau _{{\rm{iden}},n}}}
\end{array}} \right] = \left\{ {\begin{array}{*{20}{c}}
{\begin{array}{*{20}{c}}
{\begin{array}{*{20}{c}}
{\left[ {\begin{array}{*{20}{c}}
1\\
0
\end{array}} \right],}\\
{\left[ {\begin{array}{*{20}{c}}
0\\
1
\end{array}} \right],}
\end{array}}&{\begin{array}{*{20}{c}}
{\begin{array}{*{20}{c}}
{s.t.}&{\exists \left( {{\cal I}_{{\rm{surf}}}^{\left( n \right)}} \right),\mathord{\setbox0=\hbox{$\exists$}
\rlap{\raise.2ex\hbox to\wd0{\hss/\hss}}\box0} \left( {{\cal I}_{{\rm{iden}}}^{\left( n \right)}} \right)}
\end{array}}\\
{\begin{array}{*{20}{c}}
{s.t.}&{\mathord{\setbox0=\hbox{$\exists$}
\rlap{\raise.2ex\hbox to\wd0{\hss/\hss}}\box0} \left( {{\cal I}_{{\rm{surf}}}^{\left( n \right)}} \right),\exists \left( {{\cal I}_{{\rm{iden}}}^{\left( n \right)}} \right)}
\end{array}}
\end{array}}
\end{array}}\\
{\begin{array}{*{20}{c}}
{\sigma \left( {{\rm{CN}}{{\rm{N}}_n}\left( {\left[ {\begin{array}{*{20}{c}}
{{{\bf{F}}_{{\rm{surf}},n}}}\\
{{{\bf{F}}_{{\rm{iden}},n}}}
\end{array}} \right]} \right)} \right),}&{others}
\end{array}}
\end{array}} \right.
\label{eq:tauboth}
\end{equation}
where ${\rm{CNN}}\left(  \cdot  \right)$ denotes the convolutional neural network operator. $\left[  \cdot  \right]$ denotes the concatenate function.

\subsection{LR$^2$ based Decoder}
\begin{figure}[htpb]
	\centering
	\includegraphics[width=1.0\linewidth]{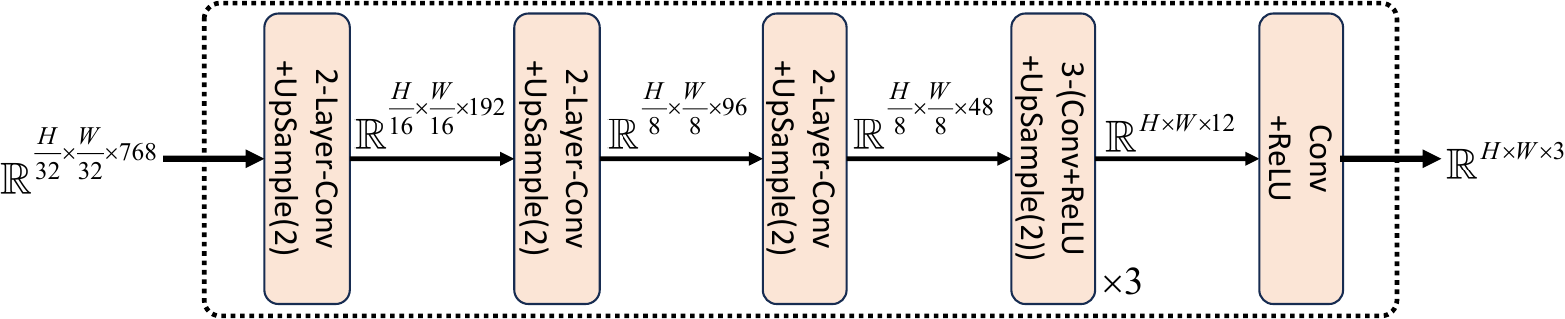}
	\caption{The inner architecture of the lightweight surface/indentation decoder.}\label{fig:decoder}
\end{figure}

The inner architecture of the LR$^2$ based decoder is shown in Fig. \ref{fig:decoder}, it contains three \{2-Layer Conv+Upsample\} blocks, three \{Conv+Upsample\} layers, and one \{Conv+ReLU\} layers. In contrast to the encoder, the decoder is designed with a lightweight architecture, which retains abundant intrinsic information embedded in feature representations by imposing reconstruction regularization.

\subsection{Optimization}
The overall loss function of the proposed algorithm is defined in Equation (\ref{eq:totalloss}).

\begin{equation}
{{\cal L}_{{\rm{total}}}} = {{\cal L}_{{\rm{cls}}}} + \alpha  \cdot {{\cal L}_{{\rm{pull}}}} + \beta  \cdot {{\cal L}_{{\rm{dis}}}}
\label{eq:totalloss}
\end{equation}
where $\alpha=0.5$ and $\beta=0.3$ are two hyperparameters to control the relative importance among different terms, while the definitions of $\mathcal L_{\rm{cls}}$, $\mathcal L_{\rm{pull}}$, and $\mathcal L_{\rm{rec}}$ are shown in Equation (\ref{eq:losscls}), Equation (\ref{eq:losspull}), and Equation (\ref{eq:lossrec}) respectively.

\begin{equation}
\begin{split}
{{\cal L}_{{\rm{cls}}}} &= \sum\limits_{n = 1}^N {{\hbar _{{\rm{cls}}}}\left( {\hat y_{{\rm{surf}}}^{\left( n \right)},{y^{\left( n \right)}}} \right) + {\hbar _{{\rm{cls}}}}\left( {\hat y_{{\rm{inde}}}^{\left( n \right)},{y^{\left( n \right)}}} \right)} \\
 &+ {\hbar _{{\rm{cls}}}}\left( {\hat y_{{\rm{fuse}}}^{\left( n \right)},{y^{\left( n \right)}}} \right)\\
\end{split}
\label{eq:losscls}
\end{equation}
where ${\hbar _{{\rm{cls}}}}$ denotes the cross entropy function. $y^{(n)}$ denotes the category label of the $n_{th}$ sample. $\hat y_{{\rm{surf}}}^{\left( n \right)}$, $\hat y_{{\rm{inde}}}^{\left( n \right)}$, and $\hat y_{{\rm{fuse}}}^{\left( n \right)}$ denote the predictions of surface branch, indentation branch and fusion branch respectively. ${{\cal L}_{{\rm{cls}}}}$ enables all branches to possess independent discriminative capacity for classification.

\begin{equation}
\begin{split}
{{\cal L}_{{\rm{pull}}}} &= \sum\limits_{n = 1}^N {{\mathchar'26\mkern-10mu\lambda _{{\rm{dis}}}}\left( {{\cal B}_{{\rm{surf}}}^{{\rm{en}}}\left( {{\cal I}_{{\rm{surf}}}^{\left( n \right)}} \right),{{\cal B}_{{\rm{fuse}}}}\left( {{\cal I}_{{\rm{surf}}}^{\left( n \right)},{\cal I}_{{\rm{inde}}}^{\left( n \right)}} \right)} \right)} \\
 &+ {\mathchar'26\mkern-10mu\lambda _{{\rm{dis}}}}\left( {{\cal B}_{{\rm{inde}}}^{{\rm{en}}}\left( {{\cal I}_{{\rm{inde}}}^{\left( n \right)}} \right),{{\cal B}_{{\rm{fuse}}}}\left( {{\cal I}_{{\rm{surf}}}^{\left( n \right)},{\cal I}_{{\rm{inde}}}^{\left( n \right)}} \right)} \right)
\end{split}
\label{eq:losspull}
\end{equation}
where ${\mathchar'26\mkern-10mu\lambda _{{\rm{dis}}}}$ denotes the KL distance function. ${\cal B}$ denotes the network of different components. $\cal I$ denotes the input image of different branches. ${\cal L}_{\rm{pull}}$ enables the surface branch and indentation branch to acquire representation capabilities comparable to those of the fusion branch, thereby achieving accurate recognition even under missing-modality conditions.

\begin{equation}
\begin{split}
{{\cal L}_{{\rm{rec}}}} &= \sum\limits_{n = 1}^N {{\ell _{{\rm{dis}}}}\left( {{\cal B}_{{\rm{surf}}}^{{\rm{de}}}\left( {{\cal I}_{{\rm{surf}}}^{\left( n \right)}} \right),{\cal I}_{{\rm{surf}}}^{\left( n \right)}} \right)} \\
 &+ {\ell _{{\rm{dis}}}}\left( {{\cal B}_{{\rm{inde}}}^{{\rm{de}}}\left( {{\cal I}_{{\rm{inde}}}^{\left( n \right)}} \right),{\cal I}_{{\rm{inde}}}^{\left( n \right)}} \right)
\end{split}
\label{eq:lossrec}
\end{equation}
where ${\ell _{{\rm{dis}}}}$ represents the Frobenius norm loss function that guides the decoder to reconstruct input images using output vectors produced by the encoder.

\section{Experiments}\label{sec:experiments}
\subsection{Datasets}
\begin{figure*}[htpb]
	\centering
	\includegraphics[width=1.0\linewidth]{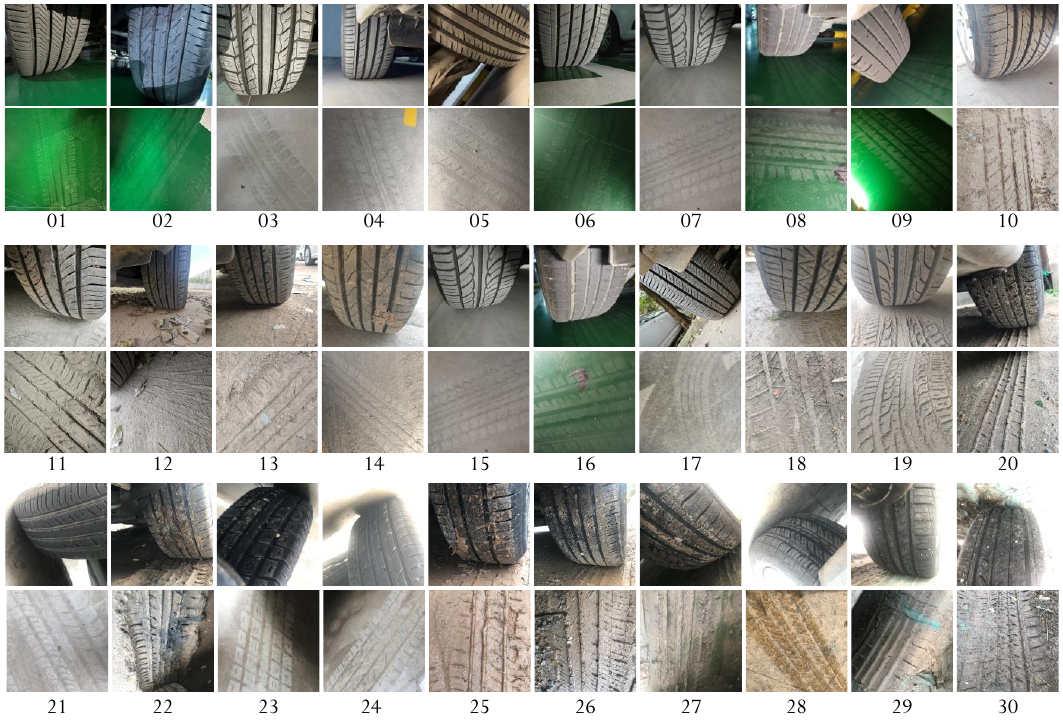}
	\caption{Selected samples of MTire299 dataset.}
        \label{fig:mtire}
\end{figure*}

\begin{figure}[htpb]
	\centering
	\includegraphics[width=1.0\linewidth]{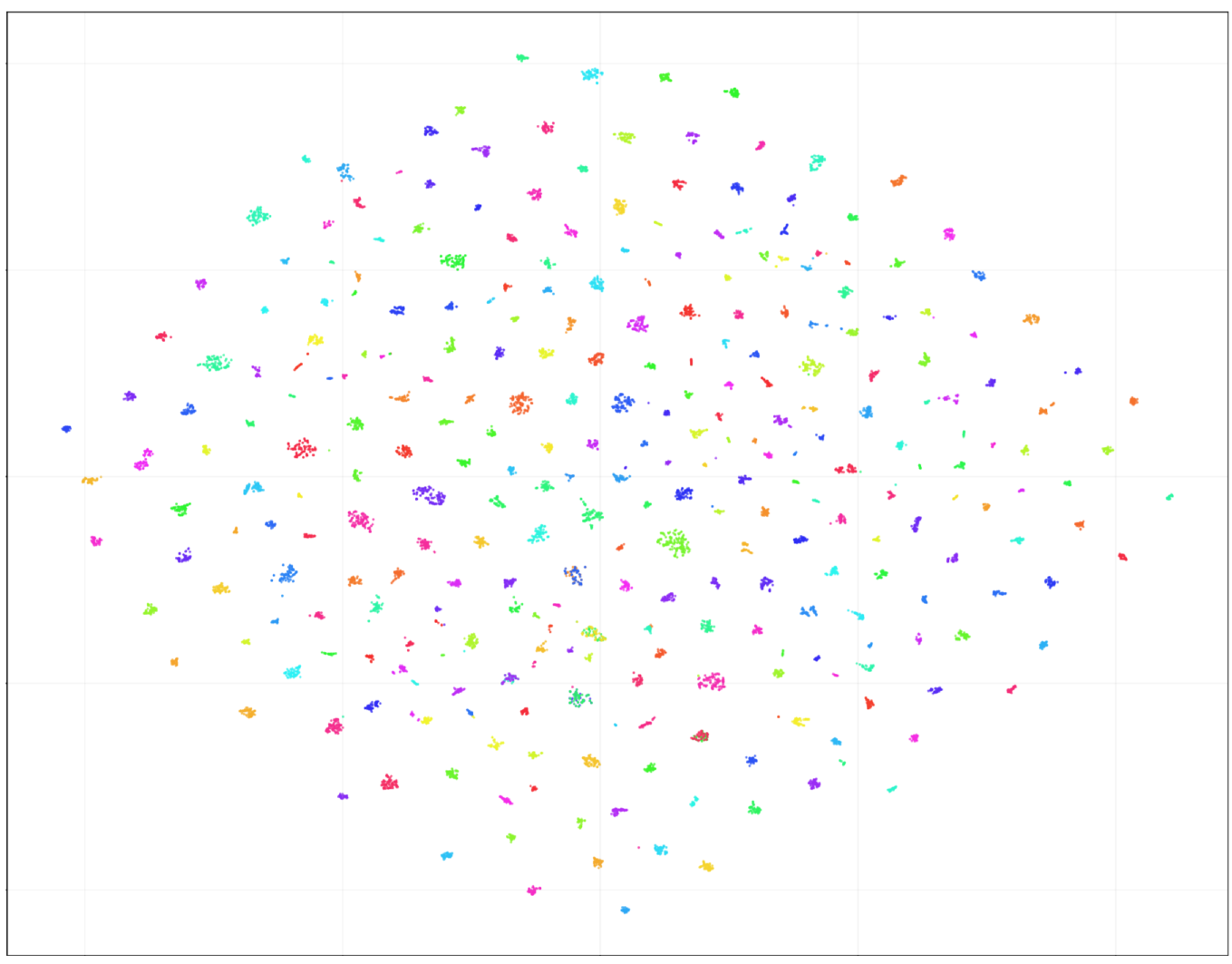}
	\caption{t-SNE visualization of fused features extracted by the proposed model on the MTire299 dataset.}
        \label{fig:tsne}
\end{figure}

We conduct a  series of experiments on two datasets, including CIIP-TPID-V1.1 \cite{hu2026cross} and MTire299. Specifically, CIIP-TPID-V1.1 \cite{hu2026cross} contains 983 images with a resolution of approximately 3000$\times$4000 pixels, covering 75 distinct tread designs across 30 brands. The self-built MTire299 contains 14795 paired surface-identation image samples within 299 fine-grained categories and each category contains between 10 and 150 paired surface-indentation samples, forming a balanced yet challenging fine-grained classification setting.and some selected samples are shown in Fig. (\ref{fig:mtire}), and the t-SNE visualization of fused features extracted by the proposed model on dataset is shown in Fig. \ref{fig:tsne}. In addition, to promote reproducibility and advance the research community of intelligent tire inspection, the complete MTire299 multi-modal dataset, covering paired surface-indentation samples, category annotations, and fixed train/val/test split lists, will be made publicly available via a dedicated open repository once this manuscript is accepted.

\subsection{Metrics}
We adopt four quantitative metrics to comprehensively evaluate model performance: classification accuracy (Acc, $\%$), parameter count (Para, in millions), GFLOPs, and frames per second (FPS).

\subsection{Comparative algorithms}
We compare our method with RiHOG \cite{liu2019rotation}, DWT-LBP \cite{liu2019wavelet}, FD  \cite{yan2021tire}, TLCNN \cite{liu2018tread}, ETL \cite{liu2018effective}, PEFT \cite{chaisiriprasert2025improving}, YOLO \cite{chuiad2024improve}, LN \cite{zhang2022lightweight}, KDAM \cite{feng2023fine}, LWLD \cite{liu2021integrating}, and UMJL \cite{zhong2024unpaired} to verify its superiority and effectiveness.

\subsection{Experimental settings}
All experiments are implemented using the PyTorch framework and executed on a single NVIDIA RTX 3080 GPU. We train the model in an end-to-end fashion with a batch size of 16. The network is optimized via the AdamW optimizer with an initial learning rate of 10$^{-3}$ and a weight decay of 10$^{-4}$. A linear learning rate scheduler is adopted throughout the training process. All training configurations are fixed for fair comparison, and random seeds are set to ensure reproducible results. For both datasets, we partitioned the data into training, validation, and test subsets following a fixed split ratio of 4 to 1 to 5.

\subsection{Experimental analysis}
\subsubsection{Contrasting experimental results}
\begin{table*}[htpb]
\centering
\caption{Experimental results on CIIP-TPID-V1.1 and MTire299 datasets.}
\renewcommand\arraystretch{1.0}
\label{tab:results}
\begin{tabular}{l|l|c|c|c|c|c|c|c|c}
\hline
\multirow{2}{*}{Category$\downarrow$}
& Datasets$\rightarrow$ &\multicolumn{4}{c|}{CIIP-TPID-V1.1}&\multicolumn{4}{c}{MTire299}\\
\cline{2-10}
& Methods$\downarrow$ & Acc.(\%) &{Para.(M)} &{GFLOPS} &{FPS (Hz)}& Acc.(\%) &{Para.(M)} &{GFLOPS} &{FPS(Hz)}\\
\hline
\multirow{2}{*}{Handcrafted}
& RiHOG \cite{liu2019rotation} &36.05 &--&--&556.34&43.21 &--&--&110.72 \\
\cline{2-10}
& DWT-LBP \cite{liu2019wavelet} &54.58 &--&--&11.19&62.69 &--&--&9.72 \\
\hline
\multirow{6}{*}{Vanilla CNNs}
& FD  \cite{yan2021tire} &78.41  &11.21&3.63&1088.31& 95.27&11.33&3.63&890.22\\
\cline{2-10}
& TLCNN \cite{liu2018tread} &84.32 &57.31&1.44&1870.09&94.26&58.23&1.44&1846.16 \\
\cline{2-10}
&ETL \cite{liu2018effective} &85.34&57.18&1.44&14.35&40.97 &58.09&1.44&4.60\\
\cline{2-10}
&PEFT \cite{chaisiriprasert2025improving} &38.09&16.03&10.49&490.78&47.61&16.32&10.49&411.38\\
\cline{2-10}
&YOLO \cite{chuiad2024improve} &74.75&36.30&24.68&347.75&61.99&36.58&24.68&332.55\\
\cline{2-10}
&LN \cite{zhang2022lightweight} &63.95&0.21&0.05&683.16&80.78 &0.27&0.05&680.85\\
\hline
\multirow{3}{*}{Sophisticated NNs}
& KDAM \cite{feng2023fine} &66.80 &2.32&0.60&541.78&82.23&2.61&0.60&533.19\\
\cline{2-10}
& LWLD \cite{liu2021integrating} &59.06 &134.57&30.94&362.15&86.98 &135.49&30.94&363.87 \\
\cline{2-10}
& UMJL \cite{zhong2024unpaired} &58.45&8.19&0.56&219.71&93.04&9.33&0.56&220.70 \\
\hline
\multirow{3}{*}{Ours}
& LR$^2$M$^2$T-M$_{\rm S}$&89.54&\multirow{3}{*}{2.0046}&\multirow{3}{*}{1.3951}&\multirow{3}{*}{386.07}&95.34&\multirow{3}{*}{2.0916} &\multirow{3}{*}{1.4120}&\multirow{3}{*}{803.59} \\
\cline{2-3} \cline{7-7}
& LR$^2$M$^2$T-M$_{\rm I}$&85.40&&&&95.93&&\\
\cline{2-3} \cline{7-7}
& LR$^2$M$^2$T-M$_{\rm {SI}}$ &94.25&&&&97.65 &&\\
\hline
\end{tabular}
\label{tab:sota_tpid}
\end{table*}
The contrasting experimental results are shown in Table \ref{tab:results}. To fully validate the overall advantages of our lightweight multi-modal framework embedded with M$^2$T cross-modal interaction and LR$^2$ reconstruction regularization, we conduct comprehensive comparative experiments covering three mainstream technical routes: handcrafted feature methods, vanilla convolutional neural networks, and sophisticated advanced deep networks. All competing approaches are trained and tested under identical data partitioning, hyperparameter configurations and hardware environments to guarantee fair and credible performance comparison. Evaluations are performed on two distinct tire datasets, including the public single-modal CIIP-TPID-V1.1 and our self-built large-scale multi-modal MTire299, to comprehensively benchmark the recognition capability and industrial deployment suitability of all methods for traffic inspection tasks.

Representative handcrafted feature approaches, namely RiHOG and DWT-LBP, produce far inferior recognition results across both datasets. On CIIP-TPID-V1.1, their highest classification accuracy only reaches 54.58$\%$, while the peak value on MTire299 merely stands at 62.69$\%$. These manually designed descriptors only capture shallow gradient and texture cues, lacking the capacity to excavate minute micro-texture discrepancies between highly analogous tire treads. Despite their minimal computation load, their weak feature discriminability cannot fulfill the high-precision identification demands of automated vehicle inspection, roadside traffic safety screening and classified waste tire recycling. Frequent misclassification severely impairs the operating efficiency of large-scale intelligent traffic detection terminals.

Transfer learning-based vanilla CNNs including FD, TLCNN and ETL attain moderate accuracy improvements against handcrafted pipelines, yet two critical defects restrict their practical usage for tire detection. FD achieves 78.41$\%$ accuracy on CIIP-TPID-V1.1 and 95.27$\%$ on MTire299, but all such networks merely accept single-modal input without cross-modal complementary fusion mechanisms. Their recognition performance degrades drastically once tire surface or indentation images are partially missing during on-site data capture. Second, basic convolutional architectures lack targeted regularization constraints, which induces severe overfitting on small annotated tire datasets and yields unstable generalization on the larger MTire299 dataset. Furthermore, TLCNN and ETL carry parameter volumes exceeding 57M alongside redundant calculations, making them incompatible with low-power, low-cost end-side inspection devices deployed widely at traffic service stations.

Advanced sophisticated neural networks integrated with attention modules, hybrid detection structures and lightweight learning strategies deliver marginal accuracy gains, but they still fail to strike a balanced performance for fine-grained tire recognition. Models such as KDAM and UMJL only execute spatial-domain feature encoding and overlook collaborative spatial-frequency modeling; their peak accuracy on MTire299 stays below 94$\%$, as they fail to extract subtle tread variations that serve as the core discriminative basis for similar tire categories. Other complex networks like LWLD stack deep layers to boost feature distinguishability, which brings parameter volume up to 134.57 M and heavy computational costs, drastically lowering inference speed and limiting deployment on resource-constrained industrial terminals.

\begin{figure*}[t]
	\centering
	\includegraphics[width=1.0\linewidth]{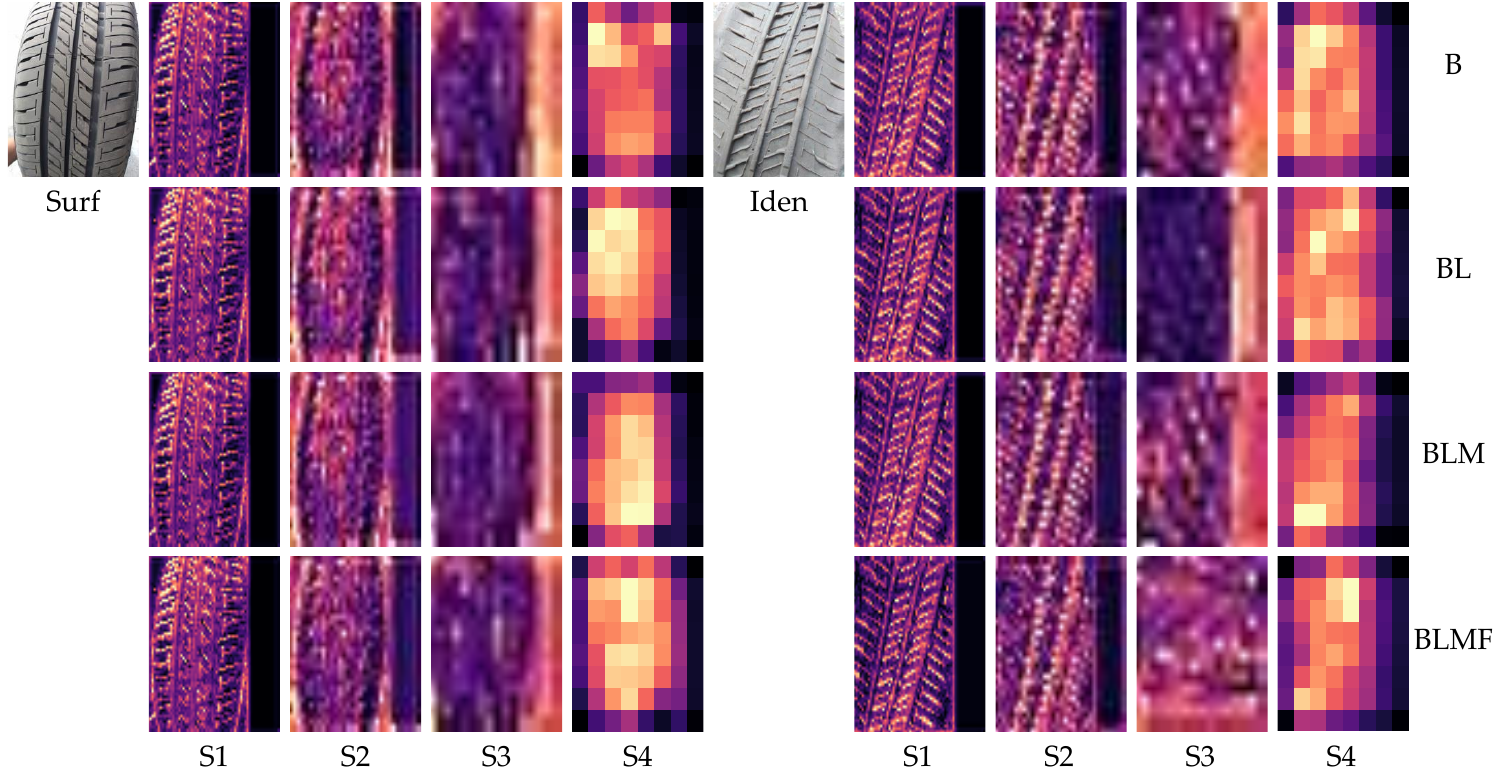}
	\caption{Feature map visualizations for four stages of the proposed method under various ablation setups. B stands for baseline without any proposed modules. BL stands for baseline + LR$^2$. BLM stands for baseline + LR$^2$ + M$^2$T. BLMF stands for full framework equipped with LR$^2$, M$^2$T and FreG. Surf and Indent correspond to feature maps of tire surface and tread indentation branches, and S1–S4 denote four successive encoder stages.}
        \label{fig:ablation}
\end{figure*}

By contrast, our full LR$^2$M$^2$T-M$_{\rm{SI}}$ framework achieves state-of-the-art recognition accuracy of 94.25$\%$ on CIIP-TPID-V1.1 and 97.65$\%$ on MTire299, outperforming all baseline methods by considerable margins. Our model maintains an extremely compact architecture with only 2.0046 M parameters and sustains fast inference speeds up to 386.07 Hz. The built-in M$^2$T module dynamically adjusts cross-branch feature weights to support stable prediction under incomplete modal inputs, perfectly adapting to the unconstrained image collection conditions of real traffic inspection sites. The introduced LR$^2$ regularization imposes reconstruction constraints on feature embeddings, effectively alleviating overfitting triggered by expensive manual annotation and limited labeled tire samples in industrial scenarios. Additionally, the hierarchical bandpass filtering unit extracts joint spatial-frequency fine-grained representations, precisely differentiating visually similar tire tread contours that single spatial encoding schemes cannot separate.

Collectively, the comparative outcomes fully verify that our method achieves an optimal trade-off between recognition precision, lightweight design and real-time inference capacity. It surpasses existing mainstream algorithms in both identification accuracy and adaptability to on-site industrial deployment, demonstrating superior practical value for intelligent traffic tire monitoring, automated vehicle inspection and multi-modal tire sorting systems.

\subsubsection{Ablation studies}
Independent ablation of individual loss terms will remove indispensable regularization and cross-modal alignment constraints, which easily causes feature divergence between the two branches and further leads to model collapse during training. Thus, we only conduct ablation studies on the proposed structural modules including the M$^2$T module, frequency-domain hierarchical guidance (FreG) unit, and LR$^2$, rather than separate loss components, on the CIIP-TPID-V1.1 dataset. All ablation variants adopt the unified ConvNeXt-Tiny dual-branch backbone and identical training configurations to ensure fair comparison. Table \ref{tab:ablationrtx4090} records quantitative metrics covering classification accuracy, parameter count, GFLOPs and inference FPS, while Fig. \ref{fig:ablation} intuitively presents feature map visualizations of four encoder stages under different ablation setups for both tire surface and tread indentation branches. 

\begin{table}[htpb]
  \centering
  \caption{Ablation study results for the CIIP-TPID-V1.1 dataset on the NVIDIA RTX 4090 platform. All ablation branches share the same backbone for equitable evaluation.}
\label{tab:ablationrtx4090}
\renewcommand\arraystretch{1.0}
\setlength{\abovecaptionskip}{1pt}
	\setlength{\belowcaptionskip}{1pt}
  \begin{tabular}{c|c|c|c|c|c|c}
    \hline
    LR$^2$ & M$^2$T &FreG& Acc. (\%) & Para. (M)&GFlops &FPS (Hz) \\
    \hline
     &                             &                              &92.81 &1.7837  &1.1677 &620.41  \\
    \hline
    $\checkmark$ &       &                               &91.94                     &1.9712 &1.3573 &409.25 \\
     \hline
    $\checkmark$      &   $\checkmark$      &&93.68                    &2.0041 &1.3615 &449.07  \\
    \hline
    $\checkmark$      &   $\checkmark$      &$\checkmark$      &94.25    &2.0046  &1.3951   &386.07    \\
    \hline
  \end{tabular}
\end{table}

The baseline model without any proposed modules achieves 92.81$\%$ accuracy with 1.7837M parameters, 1.1677 GFlops and 620.41Hz FPS. As shown by the feature maps labeled B in Fig. \ref{fig:ablation}, the plain dual-branch structure only captures rough tire contour information, and subtle discriminative tread textures are severely suppressed in deep layers, lacking cross-modal complementary interaction and frequency-aware feature modulation, which leads to weak feature discrimination and severe overfitting under limited labeled tire samples. After introducing only LR$^2$ into the baseline, the accuracy slightly drops to 91.94$\%$, accompanied by moderately increased parameters (1.9712 M), GFLOPs (1.3573) and reduced inference speed (409.25 Hz). The BL group feature maps in Fig.7 demonstrate that the reconstruction loss of LR$^2$ effectively retains inherent texture details inside feature embeddings and strengthens the response of tiny tread patterns, fundamentally alleviating overfitting risks on small-scale tire datasets despite minor accuracy loss caused by the additional reconstruction optimization objective. Further integrating the M$^2$T cross-modal interaction module on the basis of LR$^2$ raises the classification accuracy to 93.68$\%$, with marginal computational overhead growth (2.0041 M parameters, 1.3615 GFlops, 449.07 FPS). From the BLM visualization group in Fig.7, the bidirectional weight adjustment of M$^2$T realizes mutual feature compensation between the two branches: when one modality suffers occlusion or missing input in real inspection scenarios, the other branch can supply complementary fine-grained texture cues, significantly improving feature robustness against incomplete modal data. Finally, incorporating the FreG frequency guidance unit to form the complete M$^2$T-LR$^2$ framework delivers the highest accuracy of 94.25$\%$, with only tiny increments in parameters (2.0046M) and GFLOPs (1.3951), maintaining a competitive real-time speed of 386.07 Hz. The BLMF feature maps in Fig. \ref{fig:ablation} clearly reflect the advantage of joint spatial-frequency modeling: the bandpass filter-based FreG module separately extracts low-frequency global outlines and high-frequency micro-tread differences, generating highly concentrated, discriminative feature responses on subtle inter-class tread regions that single spatial encoding cannot capture. Overall, both the quantitative data in Table II and qualitative feature visualization in Fig. \ref{fig:ablation} consistently prove that LR$^2$, M$^2$T and FreG bring progressive performance gains with negligible lightweight overhead, where LR$^2$ suppresses overfitting for scarce training data, M$^2$T enables reliable prediction under missing-modality conditions, and FreG excavates minute tread distinctions for fine-grained tire classification. The full combination of the three modules achieves an optimal trade-off among recognition precision, model size and edge deployment efficiency for vehicle inspection terminals with limited computing resources.

\subsubsection{Experiments on edge devices}
To verify the deployment convenience of the proposed method on edge terminals, we deploy the well-trained models on the RK3588 computing board, and the corresponding experimental results are presented in Table \ref{tab:ablationrk3588}.

\begin{table}[htpb]
  \centering
  \caption{Ablation study results for the CIIP-TPID-V1.1 dataset on the RK3588 platform, all ablation branches share the same backbone for equitable evaluation. (INT8, 4 threads)}
\label{tab:ablationrk3588}
\renewcommand\arraystretch{1.0}
\setlength{\abovecaptionskip}{1pt}
	\setlength{\belowcaptionskip}{1pt}
  \begin{tabular}{c|c|c|c|c|c|c}
    \hline
    LR$^2$ & M$^2$T &FreG& Acc.(\%) & FPS(Hz)&NPU(\%) &Memory(MB) \\
    \hline
     &                             &                            &91.93 &92.28  &59.25 &57.29  \\
    \hline
    $\checkmark$ &       &                          &91.28 & 91.63     &71.19 &57.87 \\
     \hline
    $\checkmark$      &   $\checkmark$    &&92.37 &89.91 &69.64 &57.40  \\
    \hline
    $\checkmark$      &   $\checkmark$      &$\checkmark$      &88.67    &83.55  &64.62   &59.69    \\
    \hline
  \end{tabular}
\end{table}

All ablation models are deployed on the RK3588 edge board with INT8 quantization and 4 CPU threads. The LR$^2$-only variant achieves 91.93\% accuracy with slight quantization degradation; integrating M$^2$T raises accuracy to 92.37\% via cross-modal texture compensation, while the full pipeline with FreG drops to 88.67\% because wide-range frequency-domain operators lose critical high-frequency tread details under INT8 truncation and receive insufficient NPU hardware acceleration. Inference throughput gradually declines from 92.28Hz to 83.55Hz due to extra CPU-NPU data transmission overhead, yet all configurations satisfy real-time inspection requirements. NPU utilization moderately increases from 5.25\% to 64.62\%: M$^2$T improves parallel encoding hardware occupancy, while FreG brings fragmented frequency calculations and unstable load, leaving spare NPU computing capacity. Memory footprint mildly grows from 57.29 MB to 59.69 MB, with extra spectrum buffers in FreG as the main contributor, and all models occupy less than 180 MB RAM to support multi-camera parallel inference. Overall, the LR$^2$+M$^2$T combination delivers balanced edge performance, and our compact 2.0046 M-parameter architecture shows great practicality for low-power on-site tire inspection terminals.

\section{Conclusion}\label{sec:conclusion}
This paper designs a lightweight dual-branch tire recognition framework integrated with Mutual Modality Trust (M$^2$T) and Lightweight Reconstruction Regularization (LR$^2$). The framework resolves three persistent limitations seen in conventional recognition schemes. These limitations include overreliance on single visual modalities, inadequate joint spatial-frequency modeling for delicate tread textures, and overfitting triggered by insufficient labeled tire samples. Two parallel ConvNeXt-Tiny encoders are adopted to extract dedicated feature representations from tire surface and tread indentation images. The embedded module dynamically adjusts cross-branch feature weights to maintain stable prediction after partial modal signals disappear. Such signal loss frequently emerges during data collection on end-side intelligent devices. A hierarchical bandpass filtering unit collaboratively excavates spatial and frequency features to capture subtle tread distinctions across different tire categories. The introduced LR$^2$preserves inherent feature semantics and relieves overfitting risks brought by limited training annotations. We also construct MTire299, a large-scale paired multi-modal tire dataset to advance follow-up research on tire visual inspection. This compact architecture offers a universal solution for multi-modal fine-grained classification under data shortage and satisfies the low-computation constraints of end-side intelligent devices. Future research will boost the robustness of the model against harsh real-world disturbances and introduce incremental few-shot learning to promote the practical deployment of industrial tire inspection systems.

\bibliographystyle{IEEEtran}
\bibliography{main}

\begin{IEEEbiographynophoto}{Jianning Yang}
is currently pursuing his Ph. D with the School of Communications and Information Engineering, Xi'an University of Posts and Telecommunications, Xi'an, China.

His research interests include Multi-view Learning, Few-shot Learning, and their applications in Computer Vision.
\end{IEEEbiographynophoto}

\begin{IEEEbiographynophoto}{Jie Fang}
is currently an associate professor with the School of Communication and Information Engineering, Xi'an University of Posts and Telecommunications, Xi'an, China, and is also a research fellow $\&$ graduate industry mentor with the School of Artificial Intelligence, Optics and Electronics (iOPEN), Northwestern Polytechnical University, Xi'an, China. 

His research interests include Computer Vision, Machine Intelligence, Distributed Computation, and AI Flow.
\end{IEEEbiographynophoto}

\begin{IEEEbiographynophoto}{Xinda Ma}
is currently pursuing his master's degree with the School of Communication and Information Engineering, Xi'an University of Posts and Telecommunications, Xi'an, China.

Her research interests include Computer Vision, Machine Learning, $\pi$-Noise, and LLMs.
\end{IEEEbiographynophoto}

\begin{IEEEbiographynophoto}{Zirui Song}
is currently pursuing his master's degree with the School of Communication and Information Engineering, Xi'an University of Posts and Telecommunications, Xi'an, China.

His research interests include Computer Vision, Machine Learning, MoR, MoE, and NAS.
\end{IEEEbiographynophoto}

\begin{IEEEbiographynophoto}{Dianwei Wang}
is currently an associate professor with the School of Communication and Information Engineering, Xi'an University of Posts and Telecommunications, Xi'an, China.

His research interests include Computer Vision, Machine Intelligence, Object Detection, and Neural Network Architecture Searching.
\end{IEEEbiographynophoto}

\begin{IEEEbiographynophoto}{Nan Wang}
is currently a lecture with the School of Artificial Intelligence, Hainan Normal University, Haikou, China. 

His research interests include Machine Learning, Computer Vision, and Remote Sensing.
\end{IEEEbiographynophoto}

\end{document}